\documentclass[10pt,journal,compsoc]{IEEEtran}
\usepackage{subfigure}
\usepackage{csquotes}
\usepackage{multirow}
\usepackage{amsmath,amssymb,amsfonts}
\usepackage{algorithmic}
\usepackage{xcolor}
\usepackage{framed,multirow}
\usepackage[colorlinks=true, linkcolor = blue]{hyperref}
\usepackage{textcomp}

\ifCLASSOPTIONcompsoc
  \usepackage[compress]{cite}
\else
  \usepackage{cite}
\fi
\ifCLASSINFOpdf
  \usepackage[pdftex]{graphicx}
\else
\fi
\usepackage{amsmath}
\usepackage{amssymb}
\begin{document}
\title{Continual Learning for Peer-to-Peer Federated Learning: A Study on Automated Brain Metastasis Identification}
\author{Yixing Huang, Christoph Bert, Stefan Fischer, Manuel Schmidt, Arnd D\"orfler, \\Andreas Maier, \IEEEmembership{Senior Member, IEEE}, Rainer Fietkau, Florian Putz
\thanks{Y. Huang, C. Bert, S. Fischer, R. Fietkau, F. Putz are with Department of Radiation Oncology, University Hospital Erlangen, Friedrich-Alexander-Universit\"at Erlangen-N\"urnberg, 91054 Erlangen, Germany. They are also with Comprehensive Cancer Center Erlangen-EMN (CCC ER-EMN), 91054 Erlangen, Germany.}
\thanks{M. Schmidt and A. D\"orfler are with Department of Neuroradiology, University Hospital Erlangen, Friedrich-Alexander-Universit\"at Erlangen-N\"urnberg, 91054 Erlangen, Germany.} 
\thanks{A. Maier is with Pattern Recognition Lab, Friedrich-Alexander-Universit\"at Erlangen-N\"urnberg, 91058 Erlangen, Germany.}
}

\IEEEtitleabstractindextext{%
\begin{abstract}
Due to data privacy constraints, data sharing among multiple centers is restricted, which impedes the development of high performance deep learning models from multicenter collaboration. Continual learning, as one approach to peer-to-peer federated learning, can promote multicenter collaboration on deep learning algorithm development by sharing intermediate models instead of training data to bypass data privacy restrictions. This work aims to investigate the feasibility of continual learning for multicenter collaboration on an exemplary application of brain metastasis identification. The DeepMedic neural network is trained for brain metastasis segmentation. 920 T1 MRI contrast enhanced volumes are split to simulate multicenter collaboration scenarios. 
A continual learning algorithm, synaptic intelligence (SI), is applied to preserve important model weights for training one center after another. In a bilateral collaboration scenario, continual learning with SI achieves a sensitivity of 0.917, and naive continual learning without SI achieves a sensitivity of 0.906, while two models trained on internal data solely without continual learning achieve sensitivity of 0.853 and 0.831 only. In a seven-center multilateral collaboration scenario, the models trained on internal datasets (100 volumes each center) without continual learning obtain a mean sensitivity value of 0.699. With single-visit continual learning (i.e., the shared model visits each center only once during training), the sensitivity is improved to 0.788 and 0.849 without SI and with SI, respectively. With iterative continual learning (i.e., the shared model revisits each center multiple times during training), the sensitivity is further improved to 0.914, which is identical to the sensitivity using mixed data for training. Our experiments demonstrate the feasibility of applying continual learning for peer-to-peer federated learning in multicenter collaboration.
\end{abstract}

\begin{IEEEkeywords}
Federated learning, continual learning, multicenter collaboration, data privacy, brain metastasis segmentation, deep learning, peer-to-peer.
\end{IEEEkeywords}
}

\maketitle

\IEEEdisplaynontitleabstractindextext

%
\IEEEpeerreviewmaketitle

\section{Introduction}
\label{sect:Intro}

With the gaining importance in radiation oncology, deep learning has achieved impressive results in various tasks \cite{sahiner2019deep}, e.g., tumor segmentation \cite{kamnitsas2017efficient}, fiducial marker detection \cite{singhrao2020generative}, and dose distribution estimation \cite{xing2020feasibility}. The performance of deep learning algorithms relies highly on the amount and quality of training data. Many research centers have the required technological and human resources as well as computation power. However, the limited access to data becomes an obstacle for them to develop deep learning algorithms independently. Therefore, multicenter collaboration among different centers (including research centers and hospitals) is always important. Due to data privacy and data management regulations, e.g., the EU medical device regulation \cite{beckers2021eu}, data sharing among multiple centers is restricted, which impedes the development of high performance deep learning models from multicenter collaboration.

\begin{figure}
\centering
\begin{minipage}[b]{0.58\linewidth}
\subfigure[Center-to-peer]{
\includegraphics[width=\linewidth]{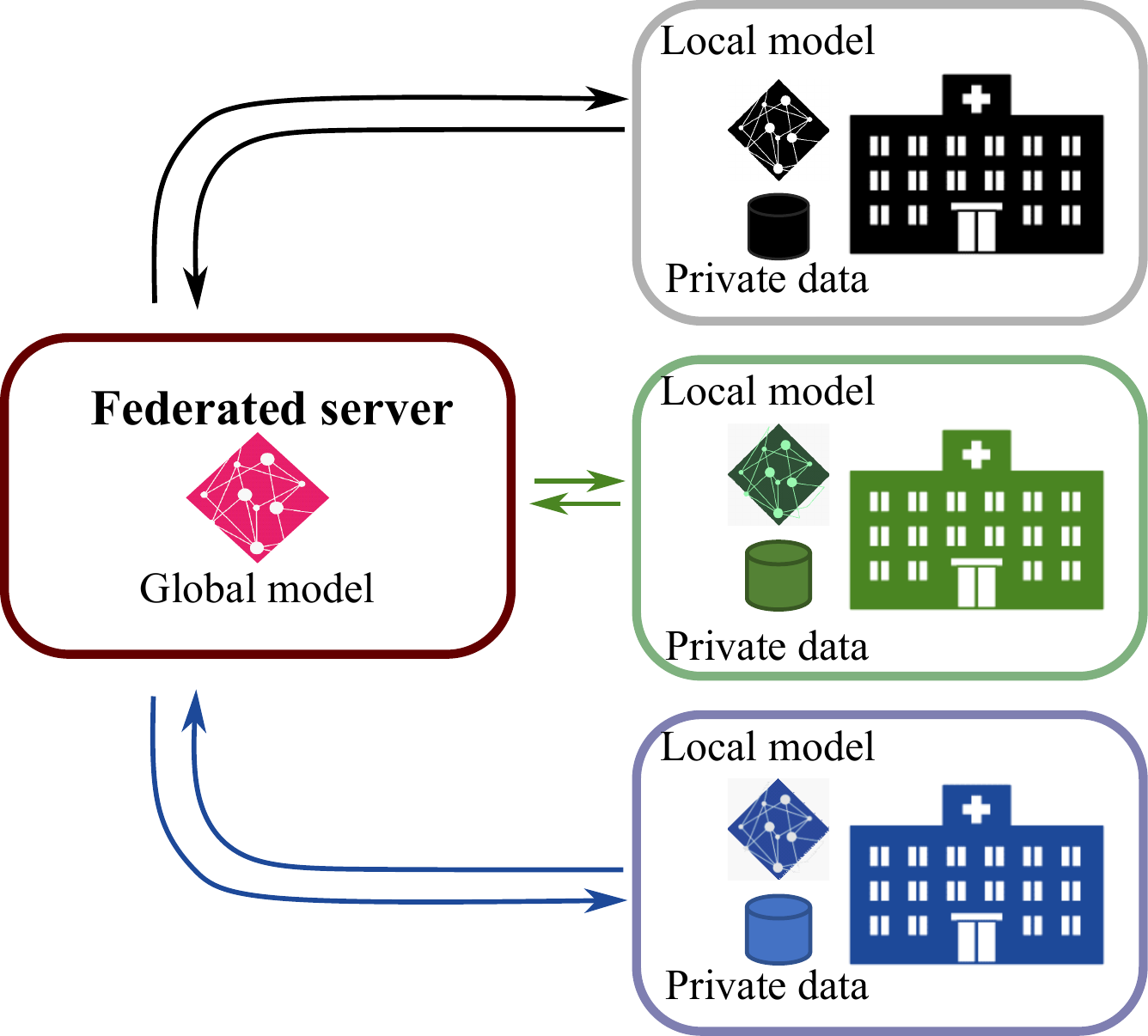}
\label{subfig:Center-to-client}
}
\end{minipage}
\begin{minipage}[b]{0.05\linewidth}
\ 
\end{minipage}
\begin{minipage}[b]{0.31\linewidth}
\subfigure[Peer-to-peer]{
\includegraphics[width=\linewidth]{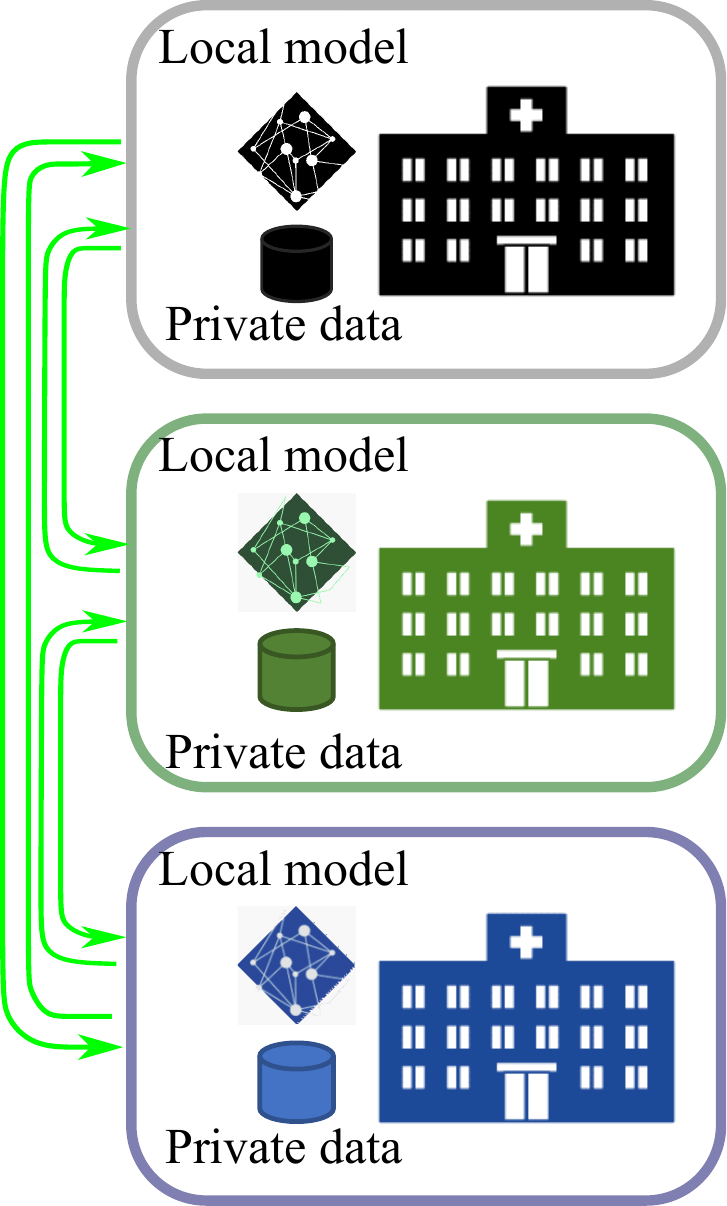}
\label{subfig:client-to-client}
}
\end{minipage}
\caption{The center-to-peer (a) and peer-to-peer (b) federated learning.}
\label{Fig:federatedLearning}
\end{figure}

To overcome the data privacy issue, federated learning has been proposed \cite{rieke2020future,sarma2021federated,xu2021federated,dayan2021federated},
 which enables multiple centers to train a high performance model without sharing data. In centralized (center-to-peer) federated learning (Fig.\,\ref{subfig:Center-to-client}), a central server is required to coordinate training information for a global model. However, such a central server is financially expensive to build and the communication cost between the central server and multiple clients is expensive as well \cite{xu2021federated}. In federated learning, reducing communication cost is one main task. For this purpose, many algorithms have been proposed. Some gradient aggregation methods, like federated averaging (FedAvg) \cite{mcmahan2017communication} and unbiased gradient aggregation \cite{yao2019federatedb}, are designed to add more computation to each client by iterating a local update and averages aggregated training gradients from all clients to save communication cost. Some utilize simplified model structures and gradient compression techniques to reduce the amount of transmitted data in each communication \cite{caldas2018expanding,konevcny2016federated,li2020federated}. Adding additional modules to each client to accelerate the convergence is another direction \cite{yao2018two,yao2019federated}. Nevertheless, the communication cost remains the main constraint for federated learning in practical applications \cite{yao2019federated,jeong2018communication}.

In IT industry, because of the extremely large number of mobile clients and the limited computational ability of mobile clients, center-to-peer federated learning is preferred. However, for multicenter collaboration in medical fields, the number of centers is typically low and each center has sufficient computation resources. Therefore, peer-to-peer (center-to-center) federated learning in a decentralised mode (Fig.\,\ref{subfig:client-to-client}) is more feasible in practice \cite{xu2021federated,wink2021approach}, since a central server is no longer required. The simplest way for such peer-to-peer federated learning is to continually train the same model one center after another via weight transfer \cite{chang2018distributed,sheller2018multi,sheller2020federated}. Chang et al. \cite{chang2018distributed} proposed distributing deep learning models via single or cyclic weight transfers for multi-institutional collaboration, which achieves comparable performance to that of centrally hosted patient data. Sheller et al. \cite{sheller2018multi,sheller2020federated} have applied the cyclic weight transfer method to glioma segmentation and confirms that cyclic weight transfer is superior to single weight transfer. However, it is inferior to center-to-peer federated learning. Especially, the performance fluctuates as forgetting \cite{mccloskey1989catastrophic,lesort2020continual,delange2021continual} occurs when the model is transferred from one institution to another. 

When a model is retrained on new datasets or tasks, deep learning suffers from the problem of catastrophic forgetting \cite{mccloskey1989catastrophic,lesort2020continual,delange2021continual}, i.e., deep learning models forget learned old knowledge catastrophically. Continual learning techniques aim to allow machine learning models updated through new data while retaining previously learned knowledge. So far, many continual learning algorithms have been proposed, which are mainly categorized to three types \cite{delange2021continual}: replay methods \cite{isele2018selective,rolnick2019experience,chaudhry2019continual}, architectural (parameter isolation) methods \cite{rusu2016progressive,fernando2017pathnet,chaudhry2019continual,mallya2018packnet} and regularization methods \cite{li2017learning,kirkpatrick2017overcoming,zenke2017continual}. Replay methods select representative samples from previous datasets to preserve learned knowledge. This is feasible to overcome data storage constraints, but not feasible for multicenter collaboration since samples from other centers are not available due to data privacy. In this case, synthetic samples from generative adversarial networks (GANs) is an alternative choice \cite{shin2017continual}. Architectural methods design dynamic network architectures or dynamic parameters for multi-task scenarios, where each part of the network (e.g., certain weights \cite{rusu2016progressive,mallya2018packnet}, or certain neuron connections \cite{fernando2017pathnet}) is responsible for each task. Regularization methods use the same conventional neural networks, but with new regularization terms in the loss function to preserve important parameters for learned knowledge, like learning without forgetting (LwF) \cite{li2017learning}, elastic weight consolidation (EWC) \cite{kirkpatrick2017overcoming} and synaptic intelligence (SI) \cite{zenke2017continual}.  

Although continual learning is known to be important in healthcare \cite{lee2020clinical}, few applications have been reported. This work applies continual learning to automatic brain metastasis (BM) identification as an exemplary case of multicenter collaboration. Patients with metastatic cancer have a high risk (an incident rate up to 40\%) of developing BM \cite{tabouret2012recent}. The early detection and successful treatment of BM are very crucial for patient survival and quality of life \cite{le2021eano}. As stereotactic radiosurgery has gained more preference than whole-brain radiotherapy for BM treatment nowadays \cite{putz2020magnetic,kocher2014stereotactic}, accurate BM identification plays a key role. Manual BM identification not only is time-consuming, but also suffers from inter/intra-rater variability. In addition, tiny metastases are easily overlooked in manual identification, since they have low contrast and appear similar to contrast enhanced blood vessels \cite{kocher2020applications}. Therefore, automatic deep learning BM identification is highly desired. Many deep learning algorithms have been proposed for this purpose, including 3D U-Net \cite{bousabarah2020deep,xue2020deep,hu2019multimodal,lu2019automated}, DeepMedic \cite{kamnitsas2017efficient,liu2017deep,lu2019automated,charron2018automatic,hu2019multimodal}, GoogLeNet \cite{grovik2020deep}, V-Net \cite{gonella2019investigating}, Faster R-CNN \cite{zhang2020deep}, and single-shot detectors \cite{zhou2020computer}. However, neither continual learning nor federated learning for automatic BM identification has been investigated.

The main contributions of this manuscript lie in the following aspects: a) To the best of our knowledge, our work is the first one to apply continual learning to address the performance fluctuation problem in cyclic weight transfer for peer-to-peer federated learning; b) Our work benchmarks the automatic BM identification performance (e.g., sensitivity and precision) over data amount, which provides hints on the amount of data each center needs to provide for such multicenter collaboration; c) Our work demonstrates the efficacy of the continual learning technique SI for single weight transfer, when the training data from each center is used once only, i.e., with minimum communication cost; d) The proposed iterative continual learning (ICL) method, i.e., cyclic weight transfer $+$ continual learning, achieves similar automatic BM identification performance compared to the optimal scenario when all the data from different centers can be shared/mixed, while the number of communications between two centers can still be kept low.

\section{Materials And Methods}

\subsection{Datasets}
In total, 920 contrast enhanced T1 volumes using the MRI magnetization-prepared rapid gradient echo (MPRAGE) \cite{brant1992mp} sequence from a longitudinal study \cite{oft2020volumetric,putz2020fsrt} are used for evaluation. MPRAGE is a standard 3D T1 inversion recovery gradient echo (IR-GRE) MRI sequence and 3D T1 IR-GRE sequences are standard sequences for BM imaging \cite{kaufmann2020consensus}. The volumes are acquired from various Siemens MRI scanners, including Aera, Avanto, TrioTim, Verio, Symphony, Espree and Sonata. The primary cancers include 41.3\% melanoma skin cancer, 22.2\% lung cancer, 12.4\% breast cancer and 10.5\% kidney cancer. On average, each volume contains 2.2 metastases. Among them, 44.4\% metastases are smaller than 0.1\,cm$^3$. All the volumes are preprocessed by the same procedures: N4ITK bias field correction \cite{tustison2010n4itk}, skull stripping \cite{isensee2019automated}, intensity normalization, volume/voxel size uniformization, and rigid transform to a template volume. 

To investigate the influence of training data amount on the BM identification performance, eight network models are trained using 100, 200, ..., 600, 700 and 750 volumes, while 67 additional volumes are used as validation data to monitor convergence and overfitting. The remaining 103 volumes are always used as test data. The volumes are split alphabetically based on patient names. In total 278 metastases exist in the test dataset. Among all the metastases, 130 (46.8\%) of them have a volume size smaller than 0.1\,cm$^3$. The volumes in the test dataset are not from the same patients in the training datasets.

\subsection{Multicenter collaboration design}

%

\begin{figure}
\centering
\includegraphics[width=\linewidth]{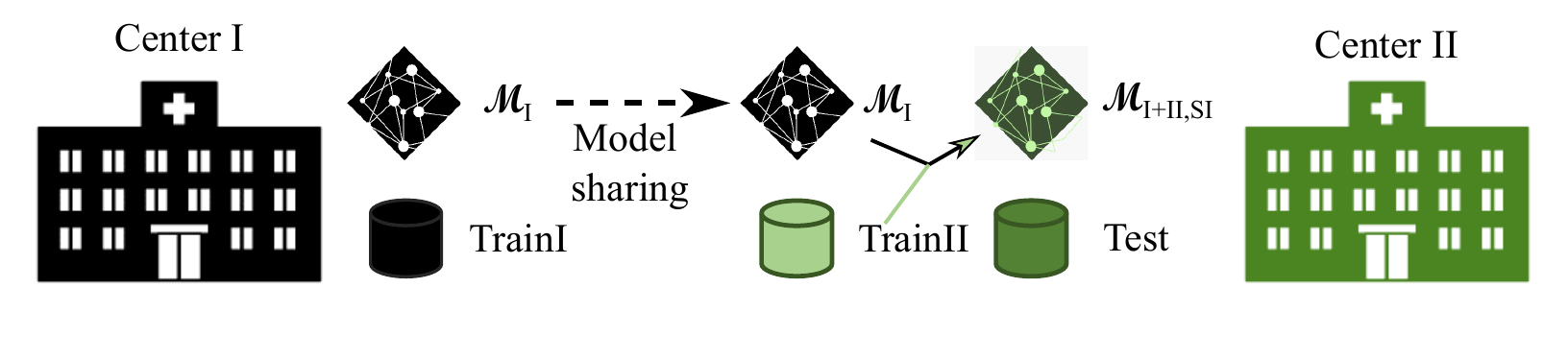}
\caption{The bilateral collaboration scheme, where a trained model $\mathcal{M}_\textrm{I}$ from center I instead of raw data TrainI is shared. The trained model $\mathcal{M}_\textrm{I}$ is trained continually from local data TrainII with SI to get a new model $\mathcal{M}_\textrm{I+II,SI}$ in center II.}
\label{subfig:twoCenterCollaboration}
\end{figure}

\color{black}
In the following, bilateral collaboration (Fig.\,\ref{subfig:twoCenterCollaboration}) and multilateral (seven-center) collaboration scenarios are simulated to investigate the benefit of continual learning in peer-to-peer federated learning.

\begin{figure*}
\centering
\subfigure[Single-visit continual learning (SVCL)]{
\includegraphics[width=0.75\linewidth]{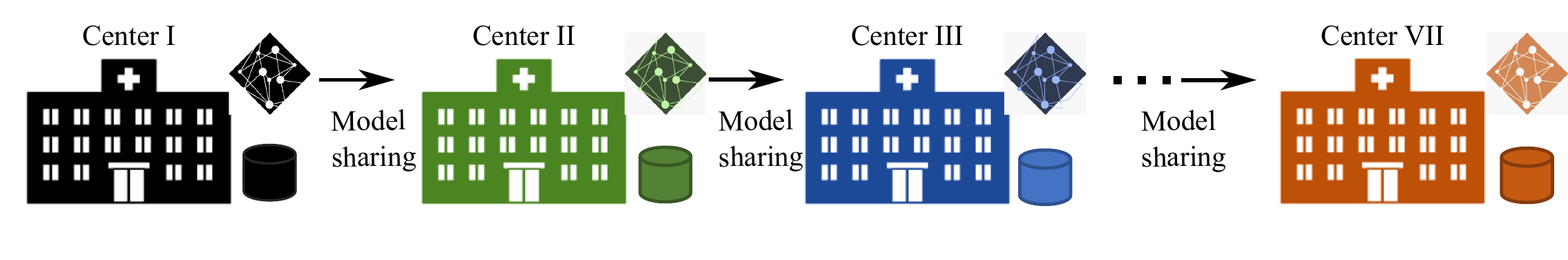}
\label{subfig:SVCL}
}

\subfigure[Iterative continual learning (ICL)]{
\includegraphics[width=0.835\linewidth]{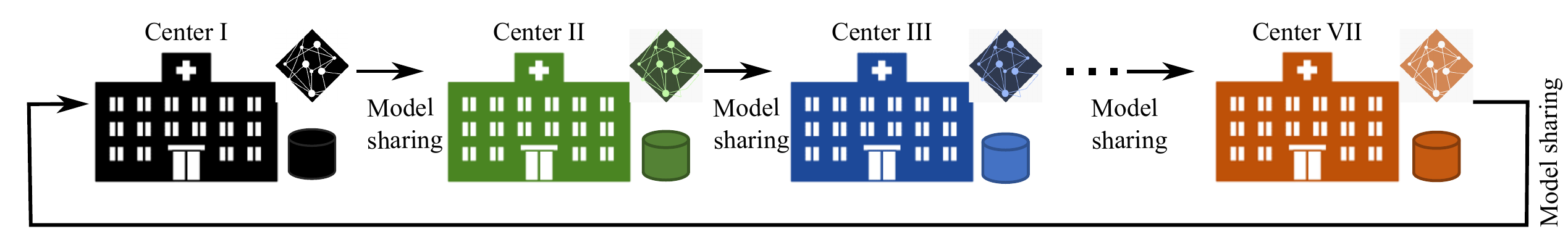}
\label{subfig:ICLscheme}
}
\caption{The multilateral (seven-center) collaboration scheme, where a trained model is shared from center I to center VII successively for continual training. The shared model visits each center once in (a), while it revisits each center iteratively in (b). The model after training at center VII is used as a final shared model for all centers.}
\label{Fig:SevenCenterCollaborationScheme}
\end{figure*}

\begin{figure*}
\centering
\includegraphics[width=0.75\linewidth]{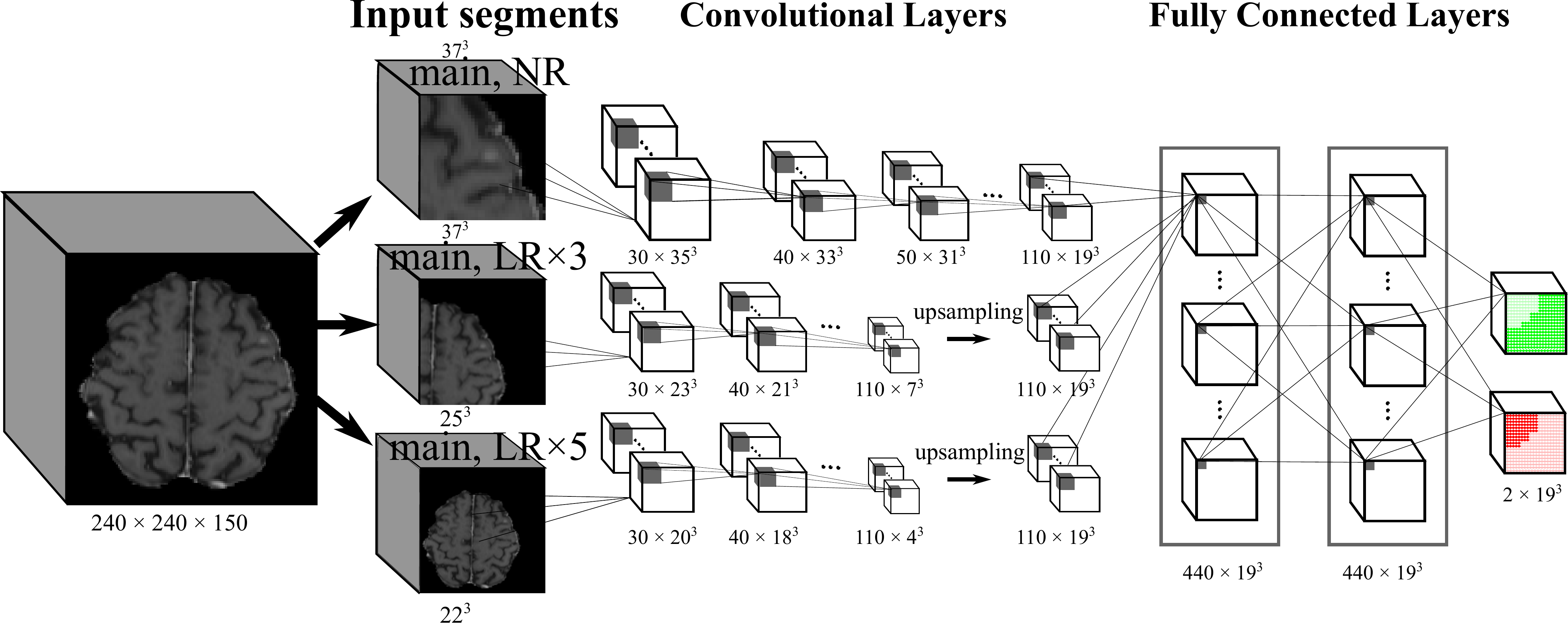}
\caption{The architecture of DeepMedic used for BM identification (figure modified from \cite{kamnitsas2017efficient}). The configuration is recommended by the authors of DeepMedic \cite{kamnitsas2017efficient}. NR and LR stand for normal resolution and low resolution, respectively.}
\label{Fig:DeepMedicArchitecture}
\end{figure*}

\subsubsection{Bilateral collaboration scenarios}
To simulate bilateral scenarios, the above 750 training volumes are equally split to two datasets: center I and center II both have 375 volumes, denoted by dataset TrainI and TrainII, respectively. Different centers do not have the volumes from the same patient. The same 67 volumes and 103 volumes are used as validation and test data respectively in center II. We aim to improve automatic BM identification performance on the test data for center II with the help of TrainI from center I.


\subsubsection{Multilateral collaboration scenario}
To simulate Multilateral (seven-center) collaboration scenarios, 700 volumes among the 750 training volumes are equally split to seven datasets, each dataset consisting of 100 volumes. The seven datasets are located in centers I to VII. The same 67 volumes and 103 volumes are used as global validation and test data, respectively. A network model is shared from center I to center VII successively for continual training, as displayed in Fig.\,\ref{Fig:SevenCenterCollaborationScheme}. The model is shared in an ordered center sequence so that each center only need to communicate with its two neighboring centers, instead of all other centers, which reduces communication cost. The model after training at center VII is used as a final shared model for all centers.

\subsection{Neural network}
The DeepMedic network \cite{kamnitsas2017efficient} (Fig.\,\ref{Fig:DeepMedicArchitecture}) is chosen because of its efficacy in various brain tumor segmentation tasks as well as its success in BM identification \cite{kamnitsas2017efficient,liu2017deep,lu2019automated,charron2018automatic,hu2019multimodal}. For BM segmentation, class imbalance is a major issue as normal tissue voxels outnumber BM voxels. To overcome this problem, DeepMedic samples volume segments online to keep class balance during training. In addition, multi-scale features are extracted via parallel convolutional pathways. As displayed in Fig.\,\ref{Fig:DeepMedicArchitecture}, features from the original scale and two coarse scales (down-sampled by factors of 3 and 5) are used. The loss function to train DeepMedic without continual learning is,
\begin{equation}
L_{\textrm{seg}} = L_{\textrm{BCE}} + L_{\textrm{VSS}},
\label{eq:SegmentationLoss}
\end{equation}
where $L_{\textrm{BCE}}$ is a conventional binary-cross entropy (BCE) loss function and $L_{\textrm{VSS}}$ is a volume-level sensitivity-specificity (VSS) loss function proposed in our previous work to improve BM sensitivity \cite{huang2021deep}. The VSS loss uses a parameter $\alpha$ to keep a trade-off between sensitivity and specificity. Since the focus of this work is on federated learning for multicenter collaboration, $\alpha$ is fixed to control its influence on model performance. 

\subsection{Continual learning}
For continual learning, architectural methods are not our choice since we aim to apply continual learning on the fixed architecture of DeepMedic. Replay methods require samples from other centers, which is typically not easily feasible due to data privacy constraints. As a result, regularization methods are of our choice. In this work, SI is investigated. SI applies a new loss function for training, which penalizes the change of important network parameters trained in previous centers. SI fundamentally has the same idea as EWC, but SI computes the importance weights during training, while EWC computes the importance weights after model training. The overall loss function $L_{\mu}$ for center $\mu$ using SI is represented as follows,
\begin{equation}
L_{\mu} = L_{\textrm{seg}} + c \sum_k \boldsymbol{\Omega}^{\mu}_k\left(\boldsymbol{\theta}^{\mu-1}_{k} - \boldsymbol{\theta}_k\right)^2,
\label{eq:continualLoss}
\end{equation}
where the left term $L_{\textrm{seg}}$ is the regular segmentation loss in Eq.\,(\ref{eq:SegmentationLoss}). The right term penalizes the change of important network parameters trained in previous centers. $c$ is the relaxation parameter to trade off old versus new knowledge. $\boldsymbol{\theta}$ is the current parameter set during training, while $\boldsymbol{\theta}^{\mu-1}$ is the parameter set trained from the previous center. $k$ is the index for a certain parameter and $\mu$ is the index for a certain center. 

The parameter importance factor $\boldsymbol{\Omega}^{\mu}$ is calculated as follows \cite{zenke2017continual},
\begin{equation}
\boldsymbol{\Omega}^{\mu}_k = \sum_{\nu < \mu} \frac{\boldsymbol{w}_k^\nu}{\left(\boldsymbol{\theta}_k^\nu - \boldsymbol{\theta}_k^{\nu-1}\right)^2 + \xi},
\label{Eqn:importanceWeight}
\end{equation}
where $\nu$ is the index of a previous center and $\boldsymbol{\theta}^\nu$ is the parameter set of the final trained model at center $\nu$. Hence, the knowledge learned from previous centers are accumulated in this importance factor. $\xi$ is a parameter to avoid division by zero. The variable $\boldsymbol{w}_k^\nu$ measures the contribution of the $k$-th network parameter to the change of the segmentation loss function $L_{\textrm{seg}}$. It is calculated as,
\begin{equation}
\boldsymbol{w}_k^\nu = \int_{t^{\nu-1}}^{t^\nu}\boldsymbol{g}_k\left(\boldsymbol{\theta}\left(t\right)\right)\cdot \boldsymbol{\theta}'\left(t\right)\textrm{d}t,
\end{equation}
where $\boldsymbol{\theta}\left(t\right)$ is the real-time network parameter set at time $t$, $\boldsymbol{g}_k = \partial \boldsymbol{L_{\textrm{seg}}}/{\partial \boldsymbol{\theta}\left(t\right)}$ is the gradient of $L_{\textrm{seg}}$ with respect to $\boldsymbol{\theta}\left(t\right)$ for the $k$-th parameter, and $\boldsymbol{\theta}'\left(t\right) = {\partial \boldsymbol{\theta}}/{\partial t}$ is the network parameter change over time.

\subsection{Iterative continual learning}
\label{subsect:ICL}
In conventional continual learning tasks, there is a typical setting that the previously accessed data is no longer accessible in the future training \cite{delange2021continual}. We call such conventional continual learning scheme ``single-visit continual learning" (SVCL), since each model visits each center only once (Fig.\,\ref{subfig:SVCL}). This is equivalent to the single weight transfer setting \cite{chang2018distributed} $+$ continual learning. However, in practical multicenter collaboration on medical applications, the training data at one center still exist after one training and hence the shared model can revisit the center for training again. We call such continual learning scheme ``iterative continual learning" (ICL), which is equivalent to the cyclic weight transfer \cite{chang2018distributed} $+$ continual learning. 
In other words, the shared model will revisit each center multiple times by sharing the model from center I to center VII iteratively in our seven-center collaboration scenario (Fig.\,\ref{subfig:ICLscheme}). 

In peer-to-peer federated learning, how often a model is shared to the next center will determine the communication cost. SVCL requires minimum communication, since two neighboring centers only need to communicate once during the training phase. As a compensation, the model performance is not optimal. For ICL, if the model is shared to the next center right after seeing all the data in the current center once (i.e., one epoch), such ICL is very close to the regular training on the mixed data. The only difference is that the training batches are fed to the network in an ordered sequence from center to center (inside each center the data is shuffled) for ICL instead of a fully randomly shuffled order. In other words, the network is using a very large ``batch" size for training, which equals the number of volumes at each center, and such large batches are fed in a certain order. Therefore, with such frequent communication (sharing the model after each epoch), the model performance in ICL should be close to that with mixed data. To reduce communication cost, a model can be shared to the next center after multiple epochs in the current center instead of one epoch only. Equivalently, multiple copies of center-level ``batches" are fed together to the network for training, which can reduce communication cost between two neighboring centers but may have performance decrease.
In our multilateral collaboration scenario, the model is shared after every 5 epochs, which can already reduce the communication number between two neighboring centers to 6 only.


\subsection{Experimental Setup}
\subsubsection{Network training parameters}
All DeepMedic models are trained on an NVIDIA Quadro RTX 8000 GPU with Intel Xeon Gold 6158R CPUs. The initial learning rate is 0.001. The learning rate decreases to half when validation accuracy plateaus. The RMSProp optimizer and Nesterov momentum \cite{kamnitsas2017efficient} with the moment value $m=0.6$, $\rho=0.9$ and $\epsilon = 10^{-4}$ are applied. Each epoch contains 20 subepochs and each subepoch uses 1000 subvolumes/segments generated randomly from 50 complete volumes for training. 
50\% probability of tumor segments are extracted as training samples to keep class-balance. The segments for the main path have a size of $37 \times 37 \times 37$. The segments are augmented with random intensity scaling, flipping and rotation. For efficient inference, $45 \times 45 \times 45$ input segments are applied. The above network configurations are recommended by the authors of DeepMedic \cite{kamnitsas2017efficient}, which can be found in their public GitHub repository\footnote{https://github.com/deepmedic/deepmedic}.
In the VSS loss (Eqn.\,(\ref{eq:SegmentationLoss})), $\alpha$ is set to 0.95, where the sensitivity stays high while the average number of false positive metastases per volume is below 1. In this work, $c$ is heuristically set to 0.1 in Eqn.\,(\ref{eq:continualLoss}). $\xi$ in Eqn.\,(\ref{Eqn:importanceWeight}) is set to $10^{-8}$. For training without continual learning, each model is trained for 70 epochs at each center and each epoch takes 20\,min (in total about 23.3 hours for one model). For training with SVCL, the model at the initial center is trained for 70 epochs as well and it is trained for 35 epochs in the following centers. For training with ICL, the model is trained for 5 epochs at each center with 6 iterations. 
With continual learning using SI, each epoch takes 22\,min (additional 2\,min because of the SI computation). As monitored by the validation data, 50 epochs and 30 epochs are sufficient for model convergence without apparent overfitting. For inference, each patient volume takes about 11\,s.

\subsubsection{Evaluation metrics}
The sensitivity, precision, and average false positive rate (AFPR) per patient volume are used for the evaluation of BM identification accuracy. Sensitivity is defined as the number of true positive BM divided by the number of true positive plus false negative BM. Precision is defined as the number of true positive BM divided by the number of true positive plus false positive BM. AFPR is defined as the average number of false positive BM per patient volume. In addition, the mean dice score coefficient (DSC) for all true positive metastases is reported. Note that only DSC is evaluated on a voxel level, while all other metrics are on a metastasis level.

\section{Results}

\subsection{Performance over training data amount}

\begin{figure}[h!]
\centering
\begin{minipage}[b]{0.8\linewidth}
\subfigure[Overall performance]{
\includegraphics[width=\linewidth]{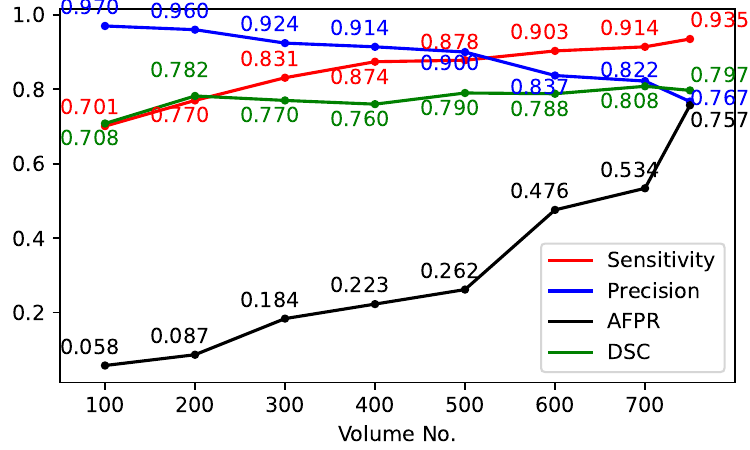}
\label{subFig:overallPerformanceOverAmount}
}
\end{minipage}

\begin{minipage}[b]{0.9\linewidth}
\subfigure[Sensitivity]{
\includegraphics[width=\linewidth]{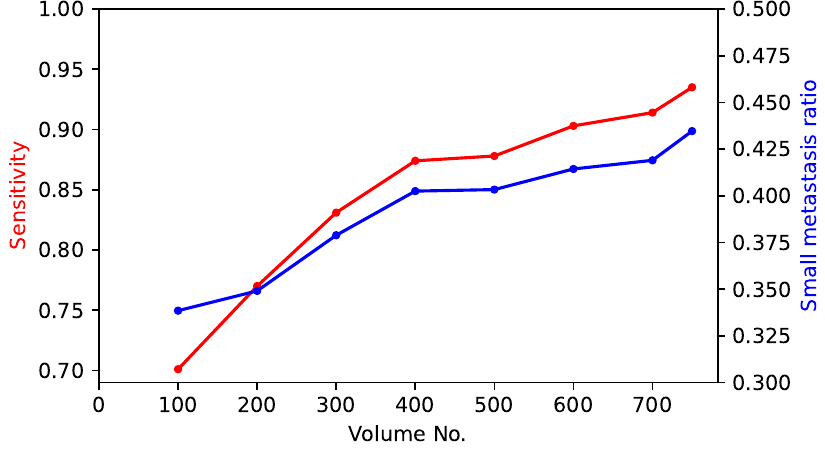}
\label{subfig:SensitivityAndRatioOverAmount}
}
\end{minipage}

%
\caption{The performance of BM identification with different numbers of training volumes: (a) overall performance with different metrics; (b) sensitivity plot as the main performance metric (red) and the ratio of small metastases among all true positive metastases (blue).}
\label{Fig:performanceOverAmount}
\end{figure}

\color{black}
The BM identification performance with different numbers of training volumes is displayed in Fig.\,\ref{subFig:overallPerformanceOverAmount}. With 100 volumes, the sensitivity, precision, AFPR and DSC are 0.701, 0.970, 0.058 and 0.708, respectively. In other words, among the overall 278 metastases, 195 metastases are correctly detected, while 83 metastases are still missing. The precision is high, with only 6 false positive metastases detected by the network. With increasing the training data amount to 750 volumes, the sensitivity and DSC values gradually increase to 0.935 and 0.797, respectively. However, the precision decreases to 0.767. Although the AFPR increases to 0.757, it is still smaller than 1 false positive metastasis per volume, which is convenient for neuroradiologists and radiation oncologists for further check in therapeutic clinical applications. Due to the relative high precision (relative low AFPR), sensitivity is used as the main performance metric of BM identification in this work. The sensitivity over various amount of training data is further plotted in a narrow range of [0.69, 1.0] in Fig.\,\ref{subfig:SensitivityAndRatioOverAmount} for a better visualization. Fig.\,\ref{subfig:SensitivityAndRatioOverAmount} indicates that with the increase of training data amount, the sensitivity is increasing in general. Fig.\,\ref{subfig:SensitivityAndRatioOverAmount} indicates that with the increase of training data amount, the ratio of small metastases (volume size $\leq 0.1\,\text{cm}^3$) among the true positive metastases is increasing as well (the reference ratio is 0.465). Therefore, more training data is beneficial to identify small metastases, which are very challenging in manual identification.

\subsection{Bilateral collaboration results}


\begin{table}[t]
\caption{BM identification performance in bilateral collaboration.}
\label{Tab:accuracyCenterII}
\centering
\begin{tabular}{l|c|c|c|c}
\hline
 Model & \textbf{Sensitivity} & Precision & AFPR & DSC\\
\hline
$\mathcal{M}_\textrm{I}$ & 0.853 &0.901 &0.252 & 0.771\\
$\mathcal{M}_{\textrm{II}}$ & 0.831 & 0.917 & 0.204 & 0.796\\
$\mathcal{M}_{\textrm{I+II,w/o SI}}$ & 0.906 &0.834 &0.485 & 0.824 \\
$\mathcal{M}_{\textrm{I+II,SI}}$ & 0.917 & 0.825 & 0.524 &0.815  \\
$\mathcal{M}_{\textrm{I+II,mix}}$ & 0.935 & 0.767 & 0.757 &0.797  \\
\hline
\end{tabular}
\end{table}

\subsubsection{Performance of different models}
\color{black}
The BM identification performance in bilateral collaboration with different models is listed in Tab.\,\ref{Tab:accuracyCenterII}. When center II reuses the trained model (denoted by $\mathcal{M}_\textrm{I}$) from center I, which is trained from TrainI solely, the sensitivity is 0.853. When center II trains a new model (denoted by $\mathcal{M}_{\textrm{II}}$) from TrainII solely, the sensitivity is 0.831. When center II continues to train $\mathcal{M}_\textrm{I}$ naively without SI (the model denoted by $\mathcal{M}_{\textrm{I+II,w/o SI}}$, which is fundamentally transfer learning, the sensitivity is 0.906. When center II continues to train $\mathcal{M}_\textrm{I}$ with SI (the model denoted by $\mathcal{M}_{\textrm{I+II,SI}}$, the sensitivity is increased to 0.917. In the ideal case when center II has the access to trainI, a model trained on the mix of trainI and trainII (denoted by $\mathcal{M}_{\textrm{I+II,mix}}$), the sensitivity is 0.935. Tab.\,\ref{Tab:accuracyCenterII} demonstrates that with continual learning naively without SI, the sensitivity is improved from 0.831 to above 0.906 for center II. With SI, the sensitivity is further improved to 0.917.

\subsubsection{Influence of parameter $c$}
 In Tab.\,\ref{Tab:accuracyCenterII}, the parameter $c$ in Eqn.\,(\ref{eq:continualLoss}) is set to 0.1. The performance of $\mathcal{M}_{\textrm{I+II,SI}}$ with two other $c$ values 0.01 and 1 has also been investigated. When $c = 0.01$, the sensitivity, precision, AFPR and DSC values are 0.874, 0.897, 0.272 and 0.758, respectively; When $c = 1$, the sensitivity, precision, AFPR and DSC values are 0.878, 0.900, 0.262 and 0.770, respectively. In both cases, the sensitivity values are lower than that (0.917) with $c=0.1$. Due to the high computation in training the 3D DeepMedic, a finer grid search for an optimal $c$ value is not performed. Therefore, in this work, we choose $c=0.1$ as a heuristic example. 


\subsection{Multilateral collaboration results}
\begin{figure}[t]
\centering
\includegraphics[width = 0.8 \linewidth]{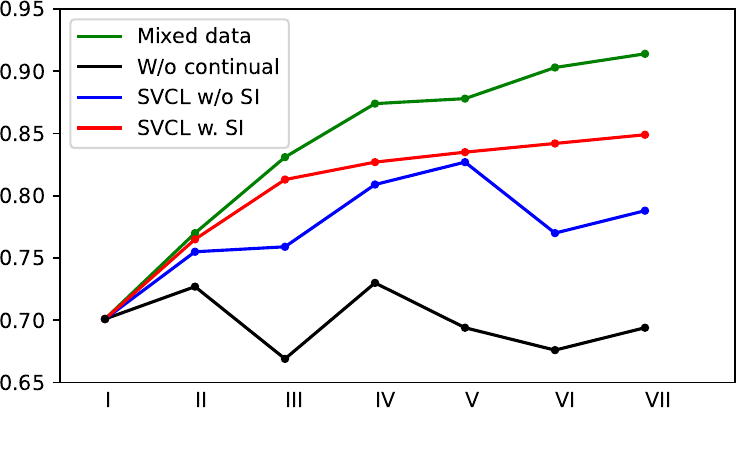}
\caption{The sensitivity plot of models trained with different strategies.}
\label{Fig:SensitivitySevenCenter}
\end{figure}

\begin{table}[t]
\caption{BM identification performance in seven-center collaboration with SVCL.}
\label{Tab:accuracySevenCenter}
\centering
\begin{tabular}{l|l|c|c|c|c}
\hline
& Center & \textbf{Sensitivity} & Precision & AFPR & DSC \\
\hline
\multirow{7}{*}{\rotatebox[origin=c]{90}{W/o continual}}&I  & 0.701 & 0.970  &0.058 & 0.708 \\
&II & 0.727 & 0.985 & 0.029 & 0.698 \\
&III  & 0.669 & 0.984 & 0.029 & 0.633\\
&IV & 0.730 & 0.962 & 0.078 &0.708 \\
&V & 0.694 & 0.937 & 0.126 &0.703 \\
&VI  & 0.676 & 0.989 & 0.019 & 0.655\\ 
&VII & 0.694 & 0.990  & 0.019 & 0.686 \\
\hline
\multirow{6}{*}{\rotatebox[origin=c]{90}{SVCL w/o SI}}&II  & 0.755 & 0.963 & 0.077 & 0.733\\
&III  & 0.759 & 0.950 & 0.107 &0.761 \\
&IV  & 0.809 & 0.904 & 0.233 &0.738 \\
&V  & 0.827 & 0.924 & 0.184 &0.761 \\
&VI  & 0.770 & 0.907 & 0.214 & 0.726\\
&VII  & 0.788 & 0.952 & 0.107 & 0.702\\
\hline
\multirow{6}{*}{\rotatebox[origin=c]{90}{SVCL w. SI}}&II & 0.766 & 0.938 & 0.136 & 0.753 \\
&III  & 0.813 & 0.934 & 0.155 & 0.754\\
&IV & 0.827 & 0.916 & 0.204 &0.749 \\
&V & 0.835 & 0.829 & 0.466 &0.741 \\
&VI & 0.842 & 0.890 & 0.282 &0.751 \\
&VII & 0.849 & 0.968 & 0.350 &0.720 \\
\hline
\end{tabular}
\end{table}

\subsubsection{Single-visit continual learning}

The sensitivity plots of different models after training in each center on the same test dataset are displayed in Fig.\,\ref{Fig:SensitivitySevenCenter}. When each center trains a model using their own data (black curve), the sensitivity values are 0.699 on average with variance from center to center. When a model is shared and continually trained from center I to center VII without SI, the sensitivity has a considerable improvement at center II and the highest sensitivity 0.827 is achieved at center V. However, a drop of sensitivity to 0.770 is observed at center VI, while it increases slightly to 0.788 at center VII. A mean sensitivity value of 0.785 is achieved. When a model is shared and continually trained from center I to center VII with SI, the sensitivity value increases smoothly one center after another. At center VII, its sensitivity is increased to 0.849. The other metrics are displayed in Tab.\,\ref{Tab:accuracySevenCenter}.

\subsubsection{Iterative continual learning}
\begin{figure*}
\centering
\includegraphics[width=1.0\linewidth]{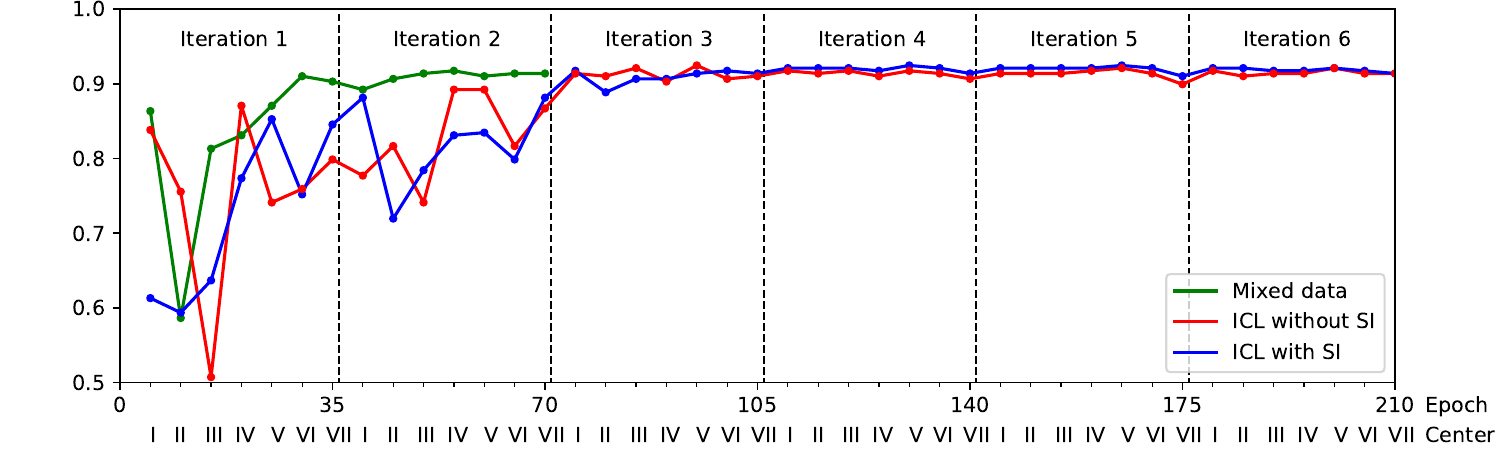}
\caption{The sensitivity plot of intermediate models for ICL without and with SI on the test dataset. The model is shared to the next center after every 5 epochs. Different iterations are concatenated for better visualization of convergence. The sensitivity plot of intermediate models trained on mixed 700 volumes is also displayed for comparison.}
\label{Fig:ICLconcat}
\end{figure*}

The sensitivity values of the shared model at different centers using different cumulative epochs are displayed in Fig.\,\ref{Fig:ICLconcat}. Note that the sensitivity is evaluated on the test dataset instead of training or validation dataset. The red curve is the sensitivity plot using naive ICL without SI. It shows that in the first two iterations, the sensitivity at different centers oscillates obviously since the model has not converged yet. At the third iteration, the sensitivity values at different centers have minor changes. From 4th to 6th iterations, the sensitivity plot changes little. 
The blue curve is the sensitivity plot using ICL with SI, which has the same trend as the red one.
For comparison, the sensitivity plot of intermediate models trained on mixed 700 volumes is also displayed (the green curve). All the three curves reach the same final sensitivity value 0.914. However, the red and blue curves need more cumulative epochs than the green one, although for each epoch all the three trainings use the same total number of $2\times 10^4$ subvolumes.
The performance metric values of the ICL models at the last iteration are displayed in Tab.\,\ref{Tab:accuracySevenCenterICL}. The sensitivity varies from 0.910 to 0.921 from center to center for ICL both without and with SI, which is from 253 to 256 true positive metastases. The precision is 0.796 and 0.789 for ICL without and with SI, respectively, slightly worse than that (0.822) trained from mixed data, which corresponds to 0.126 ($0.660-0.534$) more false positive metastases per patient or 13 ($0.126\times 103$) more false positive metastases in total.

\begin{table}[t]
\caption{BM identification performance in seven-center collaboration with ICL (last iteration).}
\label{Tab:accuracySevenCenterICL}
\centering
\begin{tabular}{l|l|c|c|c|c}
\hline
& Center & \textbf{Sensitivity} & Precision & AFPR & DSC \\
\hline
\multirow{7}{*}{\rotatebox[origin=c]{90}{ICL w/o SI}}&I & 0.917 & 0.718  &0.971 & 0.833\\
&II & 0.910 & 0.746 & 0.835 & 0.830 \\
&III  & 0.914 & 0.777 & 0.709 & 0.826\\
&IV & 0.914 & 0.779 & 0.699 &0.825 \\
&V & 0.921 & 0.736 & 0.893 &0.827 \\
&VI  & 0.914 & 0.770 & 0.738 & 0.827\\ 
&VII & 0.914 & 0.796 & 0.631 & 0.827 \\
\hline
\multirow{6}{*}{\rotatebox[origin=c]{90}{ICL w. SI}}&I  & 0.921 & 0.725 & 0.942 & 0.831\\
&II  & 0.921 & 0.757 & 0.796 &0.826 \\
&III  & 0.917 & 0.748 & 0.835 &0.824 \\
&IV  & 0.917 & 0.766 & 0.757 &0.823 \\
&V  & 0.921 & 0.751 & 0.825 &0.827 \\
&VI  & 0.917 & 0.761 & 0.777 & 0.828\\
&VII  & 0.914 & 0.789 & 0.660 & 0.821\\
\hline
\end{tabular}
\end{table}

\subsubsection{BM identification examples}

\begin{figure*}
\centering
\begin{minipage}[c]{0.22\linewidth}
\subfigure[Whole slice]{
\includegraphics[width=\linewidth]{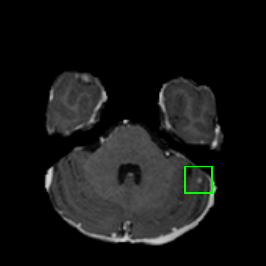}
\label{subfig:patient1}
}
\end{minipage}
\begin{minipage}[c]{0.76\linewidth}
\raggedright
\subfigure[ROI anatomy]{
\includegraphics[width=0.2\linewidth]{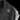}
}
\subfigure[Reference label]{
\includegraphics[width=0.2\linewidth]{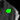}
}
\subfigure[W/o continual]{
\includegraphics[width=0.2\linewidth]{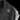}
}
\subfigure[SVCL w/o SI]{
\includegraphics[width=0.2\linewidth]{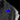}
}

\subfigure[SVCL w. SI]{
\includegraphics[width=0.2\linewidth]{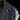}
}
\subfigure[ICL w/o SI]{
\includegraphics[width=0.2\linewidth]{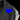}
}
\subfigure[ICL w. SI]{
\includegraphics[width=0.2\linewidth]{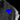}
}
\end{minipage}

\begin{minipage}[c]{0.22\linewidth}
\subfigure[Whole slice]{
\includegraphics[width=\linewidth]{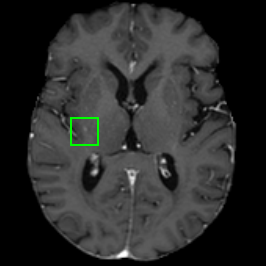}
\label{subfig:patient2}
}
\end{minipage}
\begin{minipage}[c]{0.76\linewidth}
\subfigure[ROI anatomy]{
\includegraphics[width=0.2\linewidth]{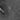}
}
\subfigure[Reference label]{
\includegraphics[width=0.2\linewidth]{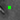}
}
\subfigure[W/o continual]{
\includegraphics[width=0.2\linewidth]{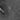}
}
\subfigure[SVCL w/o SI]{
\includegraphics[width=0.2\linewidth]{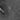}
}

\subfigure[SVCL w. SI]{
\includegraphics[width=0.2\linewidth]{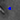}
}
\subfigure[ICL w/o SI]{
\includegraphics[width=0.2\linewidth]{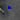}
}
\subfigure[ICL w. SI]{
\includegraphics[width=0.2\linewidth]{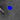}
}
\end{minipage}

\begin{minipage}[c]{0.22\linewidth}
\subfigure[Whole slice]{
\includegraphics[width=\linewidth]{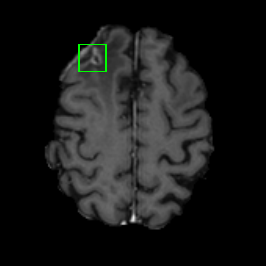}
\label{subfig:patient3}
}
\end{minipage}
\begin{minipage}[c]{0.76\linewidth}
\subfigure[ROI anatomy]{
\includegraphics[width=0.2\linewidth]{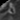}
}
\subfigure[Reference label]{
\includegraphics[width=0.2\linewidth]{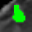}
}
\subfigure[W/o continual]{
\includegraphics[width=0.2\linewidth]{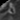}
}
\subfigure[SVCL w/o SI]{
\includegraphics[width=0.2\linewidth]{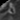}
}

\subfigure[SVCL w. SI]{
\includegraphics[width=0.2\linewidth]{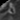}
}
\subfigure[ICL w/o SI]{
\includegraphics[width=0.2\linewidth]{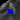}
}
\subfigure[ICL w. SI]{
\includegraphics[width=0.2\linewidth]{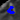}
}
\end{minipage}
\caption{The BM identification results of three exemplary patients using different methods in the seven-center multilateral collaboration scenario. The green box marks a region of interest (ROI) containing a metastasis. The green areas mark the reference manual labels, while the blue areas mark the segmented regions (true positive) by different methods. The ROIs without blue regions stand for false negative cases.}
\label{Fig:segmentationExamples}
\end{figure*}

\begin{figure*}
\centering
\begin{minipage}[c]{0.22\linewidth}
\subfigure[Whole slice]{
\includegraphics[width=\linewidth]{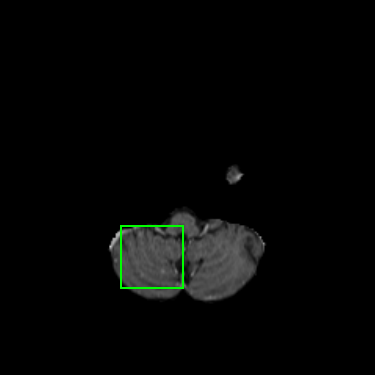}
\label{subfig:patient4}
}
\end{minipage}
\begin{minipage}[c]{0.76\linewidth}
\subfigure[ROI anatomy]{
\includegraphics[width=0.2\linewidth]{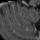}
}
\subfigure[Reference label]{
\includegraphics[width=0.2\linewidth]{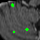}
}
\subfigure[W/o continual]{
\includegraphics[width=0.2\linewidth]{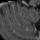}
}
\subfigure[SVCL w/o SI]{
\includegraphics[width=0.2\linewidth]{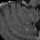}
}

\subfigure[SVCL w. SI]{
\includegraphics[width=0.2\linewidth]{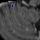}
}
\subfigure[ICL w/o SI]{
\includegraphics[width=0.2\linewidth]{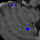}
}
\subfigure[ICL w. SI]{
\includegraphics[width=0.2\linewidth]{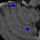}
}
\end{minipage}

%

\begin{minipage}[c]{0.22\linewidth}
\subfigure[Whole slice]{
\includegraphics[width=\linewidth]{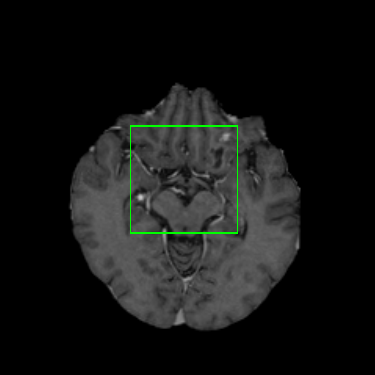}
\label{subfig:patient4}
}
\end{minipage}
\begin{minipage}[c]{0.76\linewidth}
\subfigure[ROI anatomy]{
\includegraphics[width=0.2\linewidth]{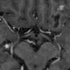}
}
\subfigure[Reference label]{
\includegraphics[width=0.2\linewidth]{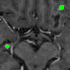}
}
\subfigure[W/o continual]{
\includegraphics[width=0.2\linewidth]{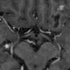}
}
\subfigure[SVCL w/o SI]{
\includegraphics[width=0.2\linewidth]{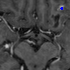}
}

\subfigure[SVCL w. SI]{
\includegraphics[width=0.2\linewidth]{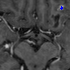}
}
\subfigure[ICL w/o SI]{
\includegraphics[width=0.2\linewidth]{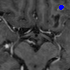}
}
\subfigure[ICL w. SI]{
\includegraphics[width=0.2\linewidth]{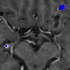}
}
\end{minipage}

%

\begin{minipage}[c]{0.22\linewidth}
\subfigure[Whole slice]{
\includegraphics[width=\linewidth]{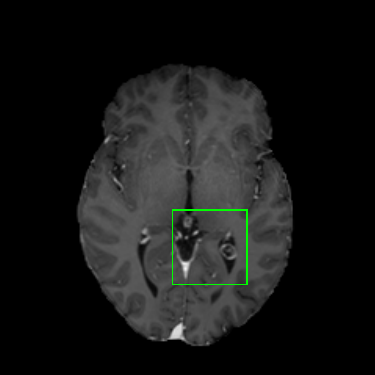}
\label{subfig:patient4}
}
\end{minipage}
\begin{minipage}[c]{0.76\linewidth}
\subfigure[ROI anatomy]{
\includegraphics[width=0.2\linewidth]{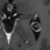}
}
\subfigure[Reference label]{
\includegraphics[width=0.2\linewidth]{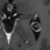}
}
\subfigure[W/o continual]{
\includegraphics[width=0.2\linewidth]{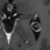}
}
\subfigure[SVCL w/o SI]{
\includegraphics[width=0.2\linewidth]{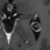}
}

\subfigure[SVCL w. SI]{
\includegraphics[width=0.2\linewidth]{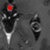}
}
\subfigure[ICL w/o SI]{
\includegraphics[width=0.2\linewidth]{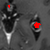}
}
\subfigure[ICL w. SI]{
\includegraphics[width=0.2\linewidth]{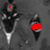}
}
\end{minipage}
\caption{The BM identification results of additional exemplary patients using different methods in the seven-center multilateral collaboration scenario. The green box marks a region of interest (ROI). The green areas mark the reference manual labels, while the blue areas mark the segmented regions (true positive) by different methods. The regions marked by green color in the reference label ROI but not marked by blue stand for false negative cases. The red regions mark false positive metastases. In (x), the left and right false positive structures are pineal gland and choroid plexus, respectively.}
\label{Fig:appendix}
\end{figure*}

The BM identification results of three exemplary patients in the seven-center multilateral collaboration scenario are displayed in Fig.\,\ref{Fig:segmentationExamples}. For the first patient (Fig.\,\ref{subfig:patient1}), DeepMedic without continual learning fails to identify the tiny metastasis, while all other methods identify it successfully. For the second patient (Fig.\,\ref{subfig:patient2}), DeepMedic without continual learning  and using native continual learning without SI both fail to identify the tiny metastasis, while the others identify it successfully. For the third patient (Fig.\,\ref{subfig:patient3}), DeepMedic using ICL both without and with SI are able to identify the tumor region, while all others fail. Due to the small sizes of these metastases (partial volume effect), it is very challenging to determine the accurate metastases' boundaries both for manual identification and DeepMedic. 
Due to limited space, false positive metastasis examples are not displayed in Fig.\,\ref{Fig:segmentationExamples}, which mainly include contrast enhanced blood vessels, choroid plexuses, and pineal glands. Three additional exemplary patients are displayed in Fig.\,\ref{Fig:appendix} as a supplementary figure.

\section{Discussion}
\color{black}
When the training data amount is small, the benefit of adding more training data is apparent, as indicated by Fig.\,\ref{Fig:performanceOverAmount}, where both the sensitivity and DSC metrics have considerable improvement without drastic decrease of precision from 100 training volumes to 200 training volumes. After reaching certain performance, adding more training data only achieves slight performance improvement only, e.g., the sensitivity increases from 0.874 (400 training volumes) to 0.914 (700 training volumes) after adding 300 training volumes more. However, with the limited data in this work, it is not sufficient to determine with how much training data the network performance will saturate. Especially, from 700 training volumes to 750 training volumes, the sensitivity increases from 0.914 to 0.935. This sensitivity improvement is relatively large, given 50 volumes are added. The network is able to identify 6 metastases more among all the 278 metastases, which has important clinical value.

In the bilateral collaboration (Tab.\,\ref{Tab:accuracyCenterII}), the sensitivity improves from 0.853 ($\mathcal{M}_\textrm{I}$) and 0.831 ($\mathcal{M}_\textrm{II}$) to 0.906 ($\mathcal{M}_{\textrm{I+II,naive}}$) and 0.917 ($\mathcal{M}_{\textrm{I+II,SI}}$) with continual learning, which is not far away from 0.935, the sensitivity of $\mathcal{M}_{\textrm{I+II,mix}}$. This demonstrates the benefit of bilateral collaboration. Bilateral collaboration is very convenient to carry out in practice and it is the starting point for multilateral collaboration. The success of two-center bilateral collaboration frequently will encourage multicenter collaboration. 

In the seven-center multilateral collaboration (Fig.\,\ref{Fig:SensitivitySevenCenter}), all the models trained on individual center dataset without continual learning achieve sensitivity values from 0.694 to 0.730, with a mean sensitivity of 0.699. Although all the individual training datasets contain 100 volumes, the model performance varies. On the one hand, deep learning networks are high dimensional and the network optimization objectives are typically nonconvex problems. Due to the existence of various local minima, training from different datasets (even with the same data amount) will lead to different local minimum solutions. On the other hand, due to the small data size at each center, the datasets have different feature distributions from each other to a certain degree. Due to the two above reasons, the sensitivity of the model without continual learning varies from center to center. Because of the small data size, the sensitivity values of all the models are below 0.75 (Tab.\,\ref{Tab:accuracySevenCenter}).

In the seven-center multilateral collaboration (Fig.\,\ref{Fig:SensitivitySevenCenter}) using SVCL without SI, the sensitivity has a considerable improvement from center I to center II and this sensitivity value 0.755 is also higher than 0.727 (the sensitivity of the model trained from individual center II data). This reveals that the sensitivity improvement from center I to center II is ascribed to not only the data variance between the two centers but also the continual learning strategy. From center III to center V, considerable sensitivity improvement is also observed.
However, a sensitivity drop is observed at center VI. This indicates that learned knowledge from previous centers are not accumulated. In other words, forgetting occurs. Nevertheless, the sensitivity values of the models with naive continual learning are all higher than those without continual learning for center II-VII, which demonstrates the benefit of continual learning.

In the seven-center multilateral collaboration (Fig.\,\ref{Fig:SensitivitySevenCenter}) using SVCL with SI, the sensitivity is increasing monotonically from center I to center VII. This highlights the efficacy of continual learning for peer-to-peer federated learning. 
The sensitivity values for continual learning with SI for all centers are higher than those with naive continual learning. This is because SI aims to preserve important network weights, which endows the network resistance to drastic performance changes (conservative), while preserving learned knowledge. Fig.\,\ref{Fig:SensitivitySevenCenter} demonstrates the importance of SI in SVCL, where communication between two neighboring centers is minimum.



In the seven-center multilateral collaboration using ICL, SI is no longer necessary as demonstrated by Fig.\,\ref{Fig:ICLconcat}. This is because the shared model revisits each center to enhance learned knowledge from that center. As described in Subsection\,\ref{subsect:ICL}, ICL fundamentally is the same as conventional learning on mixed data, but with a very large ``batch" containing multiple copies of data from each center. Using ICL, the final model performance is very close to that trained from mixed data, with identical sensitivity but higher overall computational cost and lightly worse precision. As centralized federated learning methods \cite{mcmahan2017communication,yao2019federatedb,caldas2018expanding,konevcny2016federated,li2020federated} ideally can approach to (but not exceed) the upper performance limit (training from mixed data) regarding BM identification accuracy and our ICL method has already almost reached the upper limit, the comparison of ICL-based peer-to-peer federated learning with centralized federated learning is not necessary in this work. But our proposed ICL-based peer-to-peer federated learning has drastically lower communication cost, since 6 communications only are needed between two neighboring centers. 
In addition, the number of collaborative centers for medical applications is much lower than that of mobile devices in IT-industrial applications, which is typically thousands to millions.
With such low communication cost, developing new software frameworks for federated learning is not necessary, since manually sharing the same preprocessing pipeline, deep learning training framework, and trained models is feasible.

In the state-of-the-art deep learning BM identification methods summarized in \cite{cho2021brain}, although high sensitivity above 0.9 has been achieved, AFPR stays higher than 9.0 for all the deep learning methods \cite{charron2018automatic,zhang2020deep,dikici2020automated}, as summarized in Tab.\, 2 of \cite{cho2021brain}. If the AFPR is lower than 1, the sensitivity is only 0.82 \cite{bousabarah2020deep}. The best performance reported so far is by Lu et al. \cite{lu2021randomized}, achieving a sensitivity of 0.957 and an AFPR of 0.5. But the method in \cite{lu2021randomized} requires 1288 patients with paired CT and MRI volumes for training. With ICL-based peer-to-peer federated learning, each center only needs to provide 100 volumes for training and the shared model can improve the sensitivity from 0.701 to 0.914, while keeping AFPR lower than 0.7, which is comparable/approaching to the record \cite{lu2021randomized}. This highlights the benefit of peer-to-peer federated learning. 


In this work, all the volumes from different simulated centers are actually acquired from one center. In practice, datasets from different centers may have different features because of scanners from different vendors, with different magnetic field intensities (Tesla), or even with different imaging sequences. Such data heterogeneity has not been investigated, which is one limitation of this work. Continual learning on heterogeneous data for multicenter collaboration needs further investigation. In general, acquiring data using the same image sequence category (e.g., the IR-GRE sequences \cite{kaufmann2020consensus}) and preprocessing data with the same pipeline (e.g., skull stripping, intensity normalization and volume/voxel size uniformization) are believed to reduce the gap among different centers.

\section{Conclusion}
Continual learning naively without SI (transfer learning) and with SI both can improve BM identification sensitivity for centers with limited amount of data. SI is effective to improve automatic BM identification for peer-to-peer federated learning, when minimum communication is desired (i.e., SVCL). ICL achieves comparable BM identification performance to that when all the data from different centers are shared/mixed, while the number of communications between two centers can still be low. In conclusion, continual learning is an effective approach to peer-to-peer federated learning, which has very important value for multicenter collaboration on deep learning applications.

\ 

\textbf{Human and Animal Research Disclosure:}
Ethical review and approval was not required for the study on human participants in accordance with the local legislation and institutional requirements. The patients/participants provided their written informed consent to participate in this study.

S

\end{document}